\documentclass[11pt]{article}
\usepackage[T1]{fontenc}
\usepackage[utf8]{inputenc}
\usepackage{lmodern}
\usepackage[margin=1in]{geometry}
\usepackage{cite}
\usepackage{amsmath,amssymb,amsfonts}
\usepackage{algorithmic}
\usepackage{graphicx}
\graphicspath{{figures/}{./}}
\usepackage{textcomp}
\usepackage{booktabs}
\usepackage{multirow}
\usepackage{array}
\usepackage{url}
\usepackage{xspace}
\usepackage{microtype}
\usepackage{siunitx}
\usepackage{enumitem}
\usepackage[hidelinks]{hyperref}

\newcommand{\method}[1]{\texttt{#1}}
\newcommand{\crasm}{CRASM\xspace}
\newcommand{\crasmgate}{CRASM-Gate\xspace}
\newcommand{\Rone}{\mathrm{R}@1}
\newcommand{\Rfive}{\mathrm{R}@5}
\def\BibTeX{{\rm B\kern-.05em{\sc i\kern-.025em b}\kern-.08em
        T\kern-.1667em\lower.7ex\hbox{E}\kern-.125emX}}
\newenvironment{keywords}{\par\noindent\textbf{Keywords: }}{\par}

\title{CRASM-Gate: Deterministic-First Constraint- and Role-Aware Semantic Mapping with Selective Model Assistance Across Heterogeneous Industrial Standards}
\author{Kabeh Mohsenzadegan$^{1}$, Vahid Tavakkoli$^{1}$, and Kyandoghere Kyamakya$^{1,2}$\\[0.5em]
\small $^{1}$Institue of Smart System Technologies, University of Klagenfurt, 9020 Klagenfurt, Austria\\
\small \texttt{kabeh.mohsenzadegan@aau.at}; \texttt{vahid.tavakkoli@aau.at}; \texttt{kyandoghere.kyamakya@aau.at}\\
\small $^{2}$Facult\'{e} Polytechnique, Universit\'{e} de Kinshasa, Kinshasa, Democratic Republic of the Congo}
\date{}

\begin{document}
\maketitle

\begin{center}
\small Corresponding author: Kabeh Mohsenzadegan (e-mail: \texttt{kabeh.mohsenzadegan@aau.at}).
\end{center}

\begin{center}
\small This work was partially supported by the EU-funded Arrowhead fPVN project, supported by the KDT Joint Undertaking under Grant Agreement No. 101111977.
\end{center}

\begin{abstract}
Industrial standards often encode the same engineering concept through incompatible hierarchies, identifiers, roles, and structural constraints, so the nearest lexical or embedding match can still be technically inadmissible. This article presents CRASM, a deterministic constraint- and role-aware semantic mapping method, and CRASM-Gate, its selectively model-assisted extension. The framework separates standard-specific canonicalization, bounded retrieval, deterministic rules, destination-versus-origin role interpretation, semantic and structural ranking, ambiguity refusal, and target validation. CRASM-Gate adds a confidence/disagreement gate that may invoke a candidate-constrained large language model, while final authority remains with deterministic validation. A controlled artifact covers six directed industrial-standard pairs, three difficulty levels, and ten configurations, yielding 14,400 sample-level decisions. With a fixed local model endpoint, CRASM-Gate reaches mean F1 0.9938 and an implemented structural-validity rate of 1.0000; deterministic CRASM reaches 0.9931 without generative-model calls; and the model-only baseline reaches 0.5347. Relative to the model-only baseline, CRASM reduces top-1 errors from 670 to 10 while exhibiting 0.0688 s/sample rather than 24.5263 s/sample observed latency. CRASM-Gate improves CRASM by one additional correct decision out of 1,440, with observed latency increasing to 15.7666 s/sample. The results support a deterministic-first interoperability architecture in which model assistance is optional, measurable, candidate-bounded, and unable to bypass structural validation.
\end{abstract}


\section{Introduction}
	\label{sec:introduction}
	Industrial digitalization increasingly requires information to cross boundaries between automation, asset management, sensor, control, and process-engineering ecosystems. Standardization and interoperability are therefore not peripheral implementation details but enabling conditions for digital transformation \cite{barbu2024}. Open Platform Communications Unified Architecture (OPC UA), the Asset Administration Shell (AAS), IEEE 1451, IEC 61499, and process-plant semantic models can describe the same physical quantity while assigning it different structural roles. A value described as ``temperature'' can be a variable, a submodel element, a transducer channel, a function-block interface, or a relation-associated process property. Correct interoperability therefore requires more than semantic similarity: it requires evidence about role, instance identity, datatype, unit, parent context, destination intent, and target-structure admissibility.
	
	This distinction is operationally important. A mapping can be linguistically convincing yet still select the wrong asset instance, confuse source provenance with the intended destination, map a measurement to an equipment object, or emit a path that is invalid for the destination representation. Model-driven integration and semantic technologies address parts of this problem \cite{petrasch2022,zeid2019,perzylo2019,ochoa2023}; ontology matchers such as LogMap and AgreementMakerLight combine lexical, structural, and logical evidence \cite{jimenezruiz2011,faria2013}; knowledge graphs provide relational context \cite{ji2021,tiddi2022}; and LLM-based ontology learning and matching introduce flexible language-level reasoning \cite{giglou2023,giglou2024llms4om,tufek2024validating}. None of these ingredients by itself defines who is allowed to accept a mapping when semantic plausibility conflicts with deterministic engineering constraints.
	
	This article introduces two deliberately separated methods. \crasm is the deterministic core: it combines bounded evidence retrieval, deterministic rules, explicit destination-versus-origin role interpretation, semantic/structural scoring, ambiguity refusal, and target validation. \crasmgate is the complete model-assisted framework: it preserves CRASM as the default decision path and adds an uncertainty/disagreement gate through which a candidate-constrained LLM may contribute evidence. The LLM is not permitted to invent arbitrary target paths and its output never bypasses deterministic validation. This authority boundary is the central architectural difference between using a language model as an optional uncertainty resolver and treating it as the owner of an industrial integration workflow.
	
	The study is organized around four research questions:
	\begin{itemize}[leftmargin=*]
		\item \textbf{RQ1:} How much mapping quality is provided by deterministic constraint- and role-aware reasoning relative to retrieval-, rule-, similarity-, and LLM-focused baselines?
		\item \textbf{RQ2:} What incremental quality does model assistance add beyond deterministic CRASM, and what observed latency accompanies that increment?
		\item \textbf{RQ3:} Does explicit origin-versus-destination interpretation resolve controlled context-conflict cases that defeat nearest-similarity approaches?
		\item \textbf{RQ4:} Which claims remain valid after an artifact-level audit of scenario semantics, validation scope, and reproducibility controls?
	\end{itemize}
	
	The paper makes seven contributions:
	\begin{itemize}[leftmargin=*]
		\item It formulates cross-standard integration as \emph{candidate-constrained semantic admissibility}: similarity proposes candidates, while role and structural constraints determine whether a candidate may be accepted.
		\item It introduces CRASM, a deterministic mapping method that distinguishes required destination context from source-side origin context and explicitly refuses low-margin interpretations.
		\item It introduces CRASM-Gate, a selective model-assistance architecture in which deterministic CRASM remains the default path and a bounded LLM is available only through an explicit confidence/disagreement gate.
		\item It places a strict JSON-Schema contract between probabilistic model output and deterministic software, requiring typed confidence, mapping type, transform semantics, rationale, and evidence before downstream validation.
		\item It evaluates the stable CRASM-Gate, CRASM, LLM-only, retrieval, rule, similarity, and semantic-graph operating points over six directed standard pairs and three controlled difficulty levels, while separating these stable comparisons from legacy compound ablations discovered during artifact hardening.
		\item It quantifies not only F1 and structural validity but also difficulty behavior, retrieval recall, confidence calibration, pair asymmetry, failure localization, and the marginal quality-versus-latency effect of model assistance.
		\item It hardens reproducibility through public CRASM/CRASM-Gate nomenclature, a stable Python/CLI facade, Docker evaluation, explicit publication-mode replication, official-source preparation, and a documented boundary between controlled accuracy and expert-validated real-world evidence.
	\end{itemize}
	
	\subsection{Relation to the Preliminary ISynKGR Study}
	The preliminary ISynKGR paper introduced a reproducible \emph{benchmark framework} for cross-standard semantic interoperability \cite{mohsenzadegan2026isynkgr}. Its principal contribution was the evaluation protocol rather than a standalone matcher, and its reported hybrid configuration reached F1 0.850 in that benchmark setting. The present article is intentionally reframed around a different scientific object: CRASM is a mapping method and CRASM-Gate is its bounded model-assisted extension. The numerical values from the two papers should not be read as a direct before/after score comparison because the benchmark construction and evaluation units differ; the journal contribution is methodological decomposition, explicit authority boundaries, role-aware reasoning, and a substantially stronger artifact contract.
	
	\begin{table*}[!t]
		\centering
		\caption{Scientific separation from the preliminary ISynKGR benchmark paper.}
		\label{tab:extension}
		\small
		\begin{tabular}{@{}p{0.18\textwidth}p{0.34\textwidth}p{0.40\textwidth}@{}}
			\toprule
			Dimension & Preliminary ISynKGR & This article \\
			\midrule
			Primary contribution & Reproducible benchmark framework & CRASM mapping method + CRASM-Gate framework \\
			Model role & Component of hybrid benchmark pipeline & Optional bounded uncertainty resolver \\
			Destination/origin roles & Not explicit & Explicit deterministic interpretation \\
			Final authority & Hybrid pipeline evaluation & Deterministic validation contract \\
			Structured generation & Lightweight/bounded use & Strict typed JSON Schema + downstream validation \\
			Experimental emphasis & Eight scenario families; benchmark comparison & Method decomposition, context-conflict difficulty, cost/quality, artifact audit \\
			Reproducibility surface & Benchmark scripts & Public CRASM API/CLI, Docker, multireplicate mode, official-source preparation \\
			\bottomrule
		\end{tabular}
	\end{table*}
	
	\section{Related Work and Positioning}
	\label{sec:related}
	\subsection{Industrial Semantic Interoperability}
	Industrial interoperability spans syntactic, structural, and semantic concerns. Model-driven approaches formalize transformations between heterogeneous enterprise data models \cite{petrasch2022}, while smart-manufacturing studies emphasize that interoperability remains a core barrier to end-to-end digitalization \cite{zeid2019}. OPC UA NodeSet ontologies have been proposed as semantic foundations for manufacturing-resource digital twins \cite{perzylo2019}. AAS has become a central abstraction for Industry 4.0 asset descriptions and business-process integration \cite{ochoa2023}. IEEE 1451 has been investigated as a mechanism for semantic interoperability at the transducer layer \cite{darocha2020}, and IEC 61499 provides a distributed function-block model for event-driven industrial control \cite{prenzel2019}. Work coupling AAS and ISO 15926 illustrates the need to bridge asset-centric and process-plant representations \cite{kim2022}.
	
	Recent work also demonstrates the value of semantic layers in cyber-physical systems. Steindl \emph{et al.} use knowledge-graph-based semantic event handling to support explainability across heterogeneous cyber-physical infrastructures \cite{steindl2024semantic}. Our problem is complementary: rather than explaining event behavior inside one semantic layer, we focus on selecting and validating element-level mappings between heterogeneous industrial representations.
	
	\subsection{Ontology Matching, Knowledge Graphs, and LLMs}
	Classical ontology matching combines lexical, structural, logical, and background-knowledge evidence. LogMap emphasizes scalable logic-based matching \cite{jimenezruiz2011}; AgreementMakerLight combines multiple matchers and repair mechanisms \cite{faria2013}. These systems are important reference points because they illustrate that mapping quality is rarely owned by a single similarity metric.
	
	Ontology learning predates current foundation-model approaches and established the need to derive explicit conceptual structures from heterogeneous information sources \cite{maedche2001}. Knowledge graphs provide relational context and have been widely studied for representation, acquisition, explainability, and downstream reasoning \cite{ji2021,tiddi2022}. Recent work increasingly treats knowledge graphs and LLMs as complementary rather than mutually exclusive representations \cite{pan2024unifying}. LLMs expand this toolset. LLMs4OL investigates ontology learning with large models \cite{giglou2023}, while LLMs4OM separates retrieval and matching and evaluates ontology matching over heterogeneous datasets \cite{giglou2024llms4om}. Tufek \emph{et al.} use LLMs to translate natural-language validation requirements into machine-actionable queries for semantic artifacts, including an OPC UA Robotics use case \cite{tufek2024validating}. These studies motivate LLM support for knowledge engineering, but also reinforce the need to distinguish language-level reasoning from the deterministic constraints of the target artifact.
	
	\subsection{Bounded Model Assistance and Structured Generation}
	Reasoning-and-action frameworks such as ReAct illustrate the value of separating intermediate reasoning, actions, observations, and revisions \cite{yao2023react}. Industrial integration, however, needs a stricter authority model: normalization, candidate construction, target validation, and final acceptance must remain attributable to explicit software components. CRASM-Gate therefore uses model assistance as a bounded evidence source rather than as autonomous workflow ownership.
	
	Structured output is necessary but insufficient. Grammar-constrained decoding can guarantee that generated output belongs to a prescribed language \cite{geng2023grammar}. CRASM-Gate uses a JSON Schema as the generation contract for mapping records and retains an independent semantic/structural validator. The two layers deliberately separate \emph{syntactic and typed compliance} from \emph{mapping admissibility}: a perfectly valid JSON object can still select the wrong industrial target.

	\subsection{Positioning Relative to Existing Mapping Paradigms}
	The literature suggests four recurring design paradigms: model-driven transformations, ontology matchers, graph-centric semantic integration, and LLM-supported knowledge engineering. These paradigms solve different parts of the interoperability problem. Model-driven transformations are strongest when source and target metamodels are explicit and stable \cite{petrasch2022}; logic-aware ontology matchers provide mature lexical and structural alignment mechanisms \cite{jimenezruiz2011,faria2013}; knowledge graphs expose relational context and provenance \cite{ji2021,tiddi2022}; and LLM-based approaches can interpret natural-language descriptions that are difficult to encode exhaustively as rules \cite{giglou2023,giglou2024llms4om}. CRASM-Gate is positioned between these families: it does not replace deterministic mappings or ontology matching with a generative model, but composes bounded semantic evidence with an explicit industrial acceptance contract.
	
	Table~\ref{tab:positioning} summarizes this positioning. The comparison is conceptual rather than a new head-to-head benchmark; external matchers are identified as important future experimental baselines rather than assigned unsupported scores.
	
	\begin{table*}[!t]
		\centering
		\caption{Conceptual positioning of CRASM/CRASM-Gate relative to representative interoperability paradigms.}
		\label{tab:positioning}
		\small
		\begin{tabular}{p{0.17\textwidth}p{0.24\textwidth}p{0.24\textwidth}p{0.27\textwidth}}
			\toprule
			Paradigm & Typical strength & Typical limitation for this task & CRASM-Gate relation \\
			\midrule
			Model-driven transformation & Explicit metamodel and transformation semantics & High engineering cost when heterogeneous standards expose different role/context conventions & Retains deterministic adapters and validation while allowing evidence fusion \\
			Ontology matching & Mature lexical, structural, and logical alignment & Usually assumes ontology entities rather than protocol-qualified industrial instance paths & Adopts multi-evidence ranking and refusal; external matchers remain future baselines \\
			Knowledge-graph integration & Relational context, provenance, explainability & Graph similarity alone does not define target-admissibility authority & Uses graph/retrieval evidence as input, not as final decision authority \\
			LLM-supported matching & Flexible language and metadata interpretation & Output can be structurally invalid, unbounded, or difficult to reproduce & Restricts model calls to a gate, a bounded candidate vocabulary, typed schema, and downstream validator \\
			CRASM & Deterministic, auditable role/context reasoning & Cannot recover information absent from candidates or metadata & Default decision path \\
			CRASM-Gate & Deterministic core plus selective language-model evidence & Requires profiling of gate utilization and external validation & Complete framework proposed in this article \\
			\bottomrule
		\end{tabular}
	\end{table*}
	
	\subsection{Cross-Standard Ambiguity Taxonomy}
	A common source of overclaiming in semantic-interoperability studies is to treat all standards as if they differed only in vocabulary. In the standards represented here, the same signal can occupy substantially different structural positions. OPC UA combines namespaces, objects, variables, references, and companion-model conventions \cite{perzylo2019}; AAS organizes information around assets, submodels, and submodel elements \cite{ochoa2023}; IEEE 1451 emphasizes transducer/channel descriptions \cite{darocha2020}; IEC 61499 expresses distributed control through resources and function-block interfaces \cite{prenzel2019}; and ISO 15926-oriented process-plant representations emphasize class/relation semantics, including mappings to asset-centric descriptions \cite{kim2022}. Table~\ref{tab:taxonomy} makes explicit which evidence the benchmark exercises.
	
	\begin{table*}[!t]
		\centering
		\caption{Representative semantic ambiguities exercised by the benchmark. The final column describes implemented checks, not full standards conformance.}
		\label{tab:taxonomy}
		\small
		\begin{tabular}{p{0.14\textwidth}p{0.20\textwidth}p{0.26\textwidth}p{0.30\textwidth}}
			\toprule
			Representation & Typical mapping unit & Ambiguity when context is ignored & Evidence used by CRASM \\
			\midrule
			OPC UA & Object/variable path & Same label under different namespace, object, or instance context & Qualified path, parent/object context, datatype, unit, instance identifiers \\
			AAS & Asset/submodel element & Same property name can occur under different assets or submodels & Asset/submodel context, element role, semantic label, datatype, unit \\
			IEEE 1451 & TEDS/channel element & Similar measurements can occupy different transducer/channel contexts & Channel identity, signal family, datatype, unit, parent context \\
			IEC 61499 & Function-block I/O & Measurement names can be confused with control/interface artifacts & Device/resource/FB path, I/O role, datatype, unit, signal family \\
			ISO 15926-style & Class/relation projection & Label similarity can ignore relation direction or process-plant context & Class/relation tokens, target path, signal family, contextual identity \\
			\bottomrule
		\end{tabular}
	\end{table*}
	
	\section{Problem Formulation and Scope}
	\label{sec:problem}
	\subsection{Candidate-Constrained Mapping}
	Let $S$ be a source standard and $T$ a target standard. For a source element $x \in S$, contextual evidence $E$, target rules $R_T$, and a bounded target candidate set $C_T=\{c_1,\ldots,c_n\}$, the system seeks a mapping
	\begin{equation}
		\hat{c}=\arg\max_{c_i\in C_T} F(x,c_i,E,R_T),
		\label{eq:decision}
	\end{equation}
	subject to a deterministic validation predicate
	\begin{equation}
		V_T(x,\hat{c},m,\tau)=1,
		\label{eq:validation}
	\end{equation}
	where $m$ is the mapping type and $\tau$ is an optional transform. A sample is counted as correct only when the selected target equals the hidden reference target and the resulting mapping passes the implemented contract.
	
	The task is explicitly \emph{candidate constrained}. The benchmark does not claim unrestricted discovery over an enterprise-wide namespace. Candidate sets contain 5, 9, and 13 target paths for easy, medium, and hard samples, respectively. Candidate order is shuffled, the gold target is not used as a positional hint, and medium/hard samples contain controlled same-signal distractors. This framing allows the experiment to isolate semantic-role and context disambiguation without conflating it with large-scale candidate retrieval over an unbounded repository.
	
	\subsection{Mapping Contract}
	Each accepted mapping contains a source path, target path, mapping type, optional transform, numeric confidence, rationale, and evidence list. Supported mapping types are \method{equivalent}, \method{approximate}, \method{label\_match}, \method{transform}, and \method{no\_match}. A transform object is only permitted for a transform mapping; \method{no\_match} requires an empty target path. Confidence is a number in $[0,1]$.
	
	This contract matters because the model interface and evaluator must agree on data types. During development, a free-form model could produce a semantically reasonable value such as \method{"confidence": "high"}, while the evaluator required a float. The final implementation resolves that interface mismatch at generation time through the schema rather than by silently mapping linguistic labels to arbitrary numeric values.
	
	\subsection{Architectural Authority Invariants}
	The framework enforces three control-flow properties that are stronger than output-format compliance. Let $\widetilde{C}_T$ denote the normalized target vocabulary admitted by the bounded candidate construction and target index. First, \emph{candidate closure} requires every accepted non-\method{no\_match} output $m$ to satisfy
	\begin{equation}
		\operatorname{target}(m)\in\widetilde{C}_T.
		\label{eq:candidateclosure}
	\end{equation}
	A model proposal outside this vocabulary cannot become an accepted mapping merely because it is fluent or highly confident. Second, \emph{validator dominance} requires every accepted output to satisfy the deterministic predicate in (\ref{eq:validation}); mapping type, target existence, datatype/unit compatibility, confidence, and ambiguity checks therefore remain downstream of model inference. Third, \emph{model non-authority} means that disabling model assistance yields CRASM rather than a degraded or alternate generative pipeline: retrieval, rules, role interpretation, ranking, refusal, repair, and validation remain executable without an LLM.
	
	These are architectural invariants, not semantic-correctness theorems. Candidate closure cannot guarantee that the correct target is present, and validator dominance cannot guarantee that an incomplete validator encodes every requirement of an industrial standard. Their value is narrower but operationally important: probabilistic inference cannot silently expand the target vocabulary or bypass the deterministic acceptance boundary. This makes model influence measurable and auditable at the exact point where it enters the decision process.
	
	\subsection{Standards Coverage and Non-Claims}
	The implementation covers AAS, OPC UA, IEEE 1451, IEC 61499, and an ISO 15926-style class/relation adapter. The ISO adapter is a lightweight projection used by the benchmark; it does not ingest the normative ISO 15926-2 schema or ISO 15926-4 reference-data library. The corresponding results are therefore evidence for the implemented projection, not ISO 15926 conformance.
	
	Likewise, AAS/OPC UA evaluation focuses on path-level and role-level admissibility with available datatype and unit evidence. It does not claim to synthesize a complete executable cross-model instance including every reference, namespace, semantic identifier, companion specification, and deployment constraint. These boundaries are central to the interpretation of the near-perfect controlled-benchmark scores.
	
	\section{CRASM and CRASM-Gate Architecture}
	\label{sec:architecture}
\begin{figure*}[!t]
		\centering
		\includegraphics[width=0.985\textwidth]{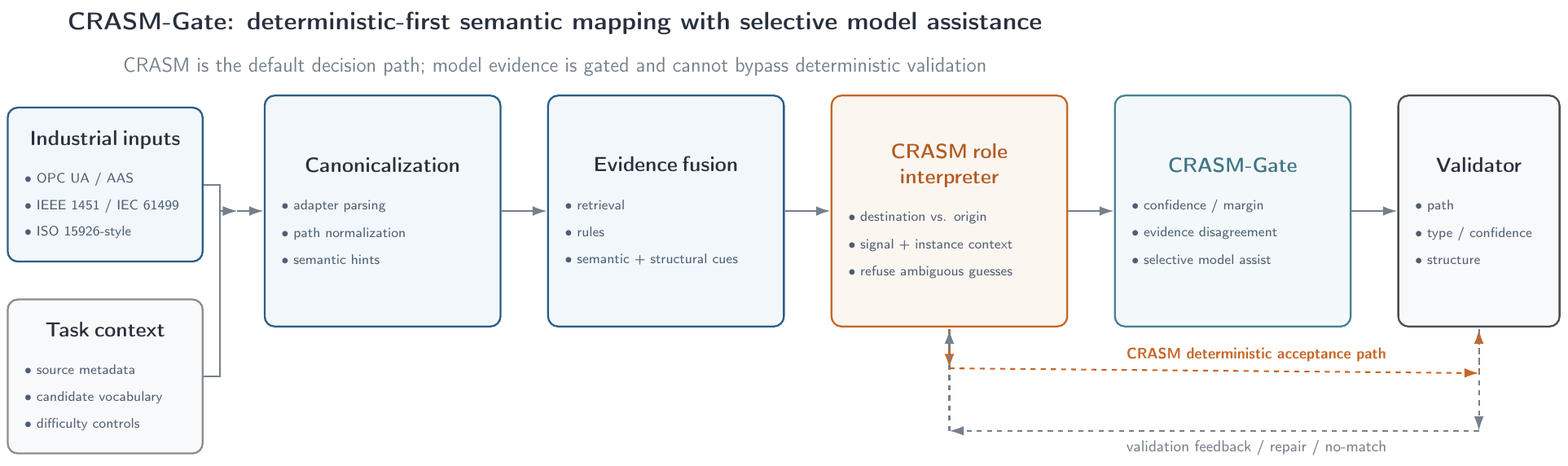}
		\caption{Vector-rendered CRASM-Gate control pipeline with orthogonal decision and feedback paths. CRASM provides the deterministic route from canonicalization through role-aware ranking and validation. CRASM-Gate can request bounded model evidence only through the gate, and every model-supported proposal returns to the same deterministic validator before acceptance.}
		\label{fig:architecture}
	\end{figure*}

	Figure~\ref{fig:architecture} separates the deterministic method from its model-assisted extension. Standard adapters normalize input elements into a canonical representation. Retrieval and rules create bounded evidence. CRASM fuses semantic, datatype, unit, instance, parent, and path cues and applies explicit destination-versus-origin role interpretation. A deterministic validator owns final acceptance. CRASM-Gate adds a confidence/disagreement decision point that can request candidate-constrained model assistance when deterministic evidence is insufficient; the model response returns to the same validation boundary.
	
	\subsection{Canonicalization and Evidence Construction}
	Standard adapters parse source artifacts and normalize identifiers into protocol-qualified paths. The system extracts labels, datatypes, units, parent paths, equipment/asset identifiers, and context tokens when available. Explicit target candidates are normalized into the same path representation. Retrieval ranks the bounded target universe; deterministic rules can add or reinforce candidates from source/target semantics. Evidence provenance is preserved so later decisions can distinguish retrieval support, rule support, contextual-role support, and model support.
	
	\subsection{CRASM Deterministic Candidate Scoring}
	CRASM uses a weighted combination of lexical similarity $L$, embedding similarity $B$, rule support $R$, datatype compatibility $D$, unit compatibility $U$, parent similarity $P$, path/context alignment $C$, retrieval score $Q$, signal-family compatibility $S$, instance alignment $I$, a measurement-role bonus $M$, and a single-candidate bonus $K$:
	\begin{align}
		S_{\mathrm{base}} ={}&0.16L+0.05B+0.18R+0.13D+0.11U \nonumber\\
		&+0.06P+0.34C+0.16Q+0.11S+0.18I \nonumber\\
		&+0.06M+0.06K.
		\label{eq:score}
	\end{align}
	The implementation then applies ambiguity, instance-conflict, origin-context, and duplicate-semantic penalties and adds a small multi-source support bonus when independent components agree. These values are version-controlled benchmark defaults, not learned parameters. Their purpose is to make the decision path inspectable and reproducible; a learned calibrator could later replace the weights without changing the mapping contract.
	
	Instance evidence is intentionally capable of overriding superficial similarity. A target with the correct signal family but a conflicting asset identifier is penalized, preventing a generic similarity score from hiding a known instance mismatch.
	
	\subsection{Destination-versus-Origin Role Interpretation}
	Controlled hard samples contain both source provenance (origin) and required target context (destination). Treating all context terms as an undifferentiated bag creates a systematic ambiguity: an origin-side decoy can appear as compatible as the intended destination. CRASM therefore extracts role-specific context tokens and scores each candidate $c$ as
	\begin{align}
		S_{\mathrm{role}}(c)={}&0.08+0.50D_c+0.16S_c+0.10T_c+0.06U_c\nonumber\\
		&+0.05M_c-0.35O_c,
		\label{eq:interpreter}
	\end{align}
	where $D_c$ is a required-destination match, $O_c$ is an origin-only match, and the remaining terms encode signal-family, datatype, unit, and measurement-role compatibility. CRASM refuses a role-level decision if the best score is below 0.76 or if the best-to-runner-up margin is below 0.04. Refusal is a first-class result: insufficient evidence returns to the broader ranker or, in CRASM-Gate, can make the case eligible for bounded model assistance.
	
	\subsection{CRASM-Gate Uncertainty and Disagreement Policy}
	Let $s_1$ and $s_2$ be the top two deterministic candidate scores. The model path is eligible when
	\begin{equation}
		G=[s_1<0.82]\lor[(s_1-s_2)\le 0.08]\lor D_{R,Q},
		\label{eq:gate}
	\end{equation}
	where $D_{R,Q}$ denotes disagreement between rule-selected and retrieval-selected targets. The thresholds are configurable and are reported here as the defaults of the evaluated implementation.
	
	The gate defines an \emph{authority boundary}, not a guarantee of sparse calls in the frozen artifact. The recorded CRASM-Gate-equivalent run marks model activity across all samples and shows heterogeneous token/cache accounting. Consequently, Section~\ref{sec:latency} treats the observed timing as a system artifact and does not claim that sparse-routing efficiency has already been demonstrated experimentally.
	
	\subsection{Strict JSON-Schema-Constrained Model Assistance}
	The model interface requires one top-level \method{mappings} array. Each element rejects additional fields and must contain source and target paths, mapping type, transform, numeric confidence, rationale, and evidence. Confidence is restricted to $[0,1]$, transform operations come from a finite enumeration, and no-match has an explicit contract.
	
	The resolved schema is passed directly as the response format and participates in the cache key, preventing stale free-form outputs from silently satisfying a newer contract. Downstream normalization remains strict: textual values such as \method{high}, \method{medium}, or \method{low} are not converted into researcher-chosen numeric scores. This protects the boundary between probabilistic inference and deterministic evaluation.
	
	\subsection{Deterministic Validation and Global Selection}
	Candidate states are rechecked for allowed mapping types, protocol-qualified paths, target existence where required, datatype compatibility, unit consistency, confidence, and ambiguity. Retrieval-only candidates below the configured confidence floor are rejected unless stronger context or instance evidence justifies acceptance. When semantically equivalent targets remain within the ambiguity margin and neither has a decisive context/instance advantage, CRASM can return no-match instead of forcing a guess.
	
	Algorithm~\ref{alg:pipeline} summarizes the control logic.
	
	\begin{figure}[!t]
		\small
		\begin{algorithmic}[1]
			\STATE Parse and normalize source model and bounded target candidates
			\STATE Retrieve candidates and apply deterministic rules
			\FOR{each source element $x$}
			\STATE Build candidate states and compute semantic/structural features
			\STATE Apply CRASM role interpretation when contextual roles are present
			\STATE Rank candidates; compute $s_1$, $s_2$, and evidence disagreement
			\IF{CRASM-Gate is enabled and condition (\ref{eq:gate}) is true}
			\STATE Request schema-constrained candidate-only model evidence
			\STATE Normalize and fuse only admissible model output
			\ENDIF
			\STATE Validate path, mapping type, unit, datatype, confidence, and ambiguity
			\STATE Accept best valid mapping, repair, or emit no-match
			\ENDFOR
			\STATE Emit mapping, provenance, trace, and validation outcome
		\end{algorithmic}
		\caption{CRASM-Gate procedure. Setting model assistance off yields deterministic CRASM.}
		\label{alg:pipeline}
	\end{figure}

	\subsection{Design Properties and Traceability Contract}
	The implementation exposes four properties that are useful for audit-sensitive deployment. \emph{Candidate boundedness} ensures that a probabilistic component cannot silently expand the target namespace. \emph{Decision decomposability} preserves distinct support channels for retrieval, rules, role interpretation, and model evidence rather than collapsing them into one opaque score. \emph{Refusal admissibility} permits a no-match result when evidence is insufficient or contradictory. \emph{Validator finality} ensures that even a model-supported candidate is subjected to the same deterministic acceptance checks as a CRASM-only candidate.
	
	These properties define what can be reconstructed after a decision. A useful mapping record should answer at least five questions: what source element was interpreted; which target candidates were available; which evidence channels supported the winner; whether model assistance was invoked and why; and which validation checks allowed or rejected the final decision. Table~\ref{tab:traceability} maps these questions to artifact fields.
	
	\begin{table}[!t]
		\centering
		\caption{Traceability questions and corresponding CRASM-Gate evidence.}
		\label{tab:traceability}
		\scriptsize
		\begin{tabular}{p{0.39\columnwidth}p{0.53\columnwidth}}
			\toprule
			Audit question & Recorded evidence \\
			\midrule
			What was mapped? & source/target protocol-qualified paths \\
			What alternatives existed? & bounded candidate list and ranking trace \\
			Why did a candidate win? & score breakdown and support sources \\
			Was a model used? & gate reason, invocation trace, model evidence \\
			Why was output accepted? & validation outcome, rejection reasons, no-match/repair trace \\
			\bottomrule
		\end{tabular}
	\end{table}
	
	This traceability contract is deliberately stronger than asking an LLM for a rationale. A generated explanation can be useful context, but it is not treated as proof of correctness. The auditable evidence is the combination of bounded candidates, deterministic feature values, component support, gate state, and validator outcome.
	
	\section{Benchmark and Experimental Protocol}
	\label{sec:benchmark}
	\subsection{Directed Standard Pairs and Difficulty}
	The frozen benchmark evaluates six directed pairs: AAS$\rightarrow$OPC UA, IEC 61499$\rightarrow$IEEE 1451, IEEE 1451$\rightarrow$IEC 61499, ISO15926-style$\rightarrow$AAS, OPC UA$\rightarrow$AAS, and OPC UA$\rightarrow$IEEE 1451. Each pair contains 240 samples divided evenly into 80 easy, 80 medium, and 80 hard cases.
	
	Easy samples expose distinct semantic or structural cues. Medium samples introduce same-signal candidates that differ in destination context. Hard samples add an origin-context decoy while preserving the intended destination in source metadata. Candidate sets contain 5, 9, and 13 entries across the three tiers. The progression is designed to test whether a method can move beyond nearest lexical or retrieval similarity.
	
	\subsection{Publication-Facing Configurations and Historical IDs}
	The result artifact was generated before the public CRASM nomenclature was introduced, so its scenario identifiers are preserved for exact reproducibility. Table~\ref{tab:names} separates publication names from frozen internal IDs.
	
	\begin{table}[!t]
		\centering
		\caption{Publication names and frozen artifact identifiers.}
		\label{tab:names}
		\scriptsize
		\begin{tabular}{ll}
			\toprule
			Publication name & Frozen ID \\
			\midrule
			CRASM-Gate & \method{full\_framework} \\
			CRASM & \method{agentic\_interpreter} \\
			LLM-only & \method{llm\_only} \\
			RAG-only & \method{rag\_only} \\
			Embedding Similarity & \method{embedding\_similarity} \\
			Semantic Graph Calibrated & \method{semantic\_graph\_calibrated} \\
			Rules-only & \method{rule\_based\_only} \\
			\bottomrule
		\end{tabular}
	\end{table}
	
	Three historical rows require stronger care. In the frozen artifact, the legacy no-rules and no-retrieval configurations did not retain the complete role-interpretation path and therefore are \emph{compound} rather than one-component ablations. Likewise, the historical no-LLM configuration removes the interpretation/feedback stack in addition to generative calls. We retain these rows for provenance and descriptive diagnostics but do not use the first two as causal estimates of the isolated contributions of rules or retrieval.
	
	The current repository corrects the no-rules and no-retrieval definitions so role-aware interpretation is retained when those components are removed. Those cleaned definitions require a fresh benchmark run before their values can replace the frozen rows in a final causal-ablation table. This distinction prevents a repository refactor from silently changing the semantics of an already reported experiment.
	
	\subsection{Metrics and Experimental Unit}
	The evaluator reports precision, recall, F1, implemented structural-validity rate, transform correctness, retrieval recall at ranks 1 and 5, confidence calibration error, latency per sample, and runtime per scenario. Each controlled sample has one reference mapping and one top decision, so precision, recall, and F1 coincide in this artifact. F1 is used as the primary quality metric while validity, retrieval, calibration, and latency remain orthogonal diagnostics.
	
	The frozen result package contains $6\times10=60$ pair/configuration records and $60\times240=14{,}400$ sample-level decisions under seed 2105616649. There are no skipped directed pairs. Standard deviations across six pair-level values describe \emph{heterogeneity among mapping directions}; they are not repeated-seed confidence intervals.
	
	\subsection{Reproducibility Controls}
	The current CRASM-Gate repository exposes publication-facing modes \method{crasm} and \method{crasm\_gate} through a public Python/CLI facade while retaining historical internal IDs only for artifact compatibility. The evaluation pipeline runs software tests and package construction before the benchmark, and Docker provides one canonical evaluation entry point. Routine evaluation defaults to a single complete run; publication mode is explicit and supports independent multi-replicate execution with separate seed streams. Machine-readable JSON/CSV metrics, logs, plots, and HTML reports are emitted as artifacts. Official OPC UA/AAS source acquisition and integrity checks are also supported, but real-world F1 remains gated on independent expert annotation; the software does not fabricate ``expert gold.''
	
	\subsection{Fixed Model Backend}
	All configurations that require generative inference use a single local Ollama backend, \method{gemma4:e2b}. This model identifier is held fixed across the frozen artifact so that differences between CRASM-Gate and LLM-only arise from the surrounding decision architecture rather than from model substitution. Deterministic configurations such as CRASM, RAG-only, Embedding Similarity, and Rules-only do not invoke the model. We therefore interpret the reported results as an evaluation of \emph{decision organization under a fixed model endpoint}, not as an optimization study over model families.
	
	\section{Results}
	\label{sec:results}
	\subsection{Aggregate Mapping Quality}
\begin{figure*}[!t]
		\centering
		\includegraphics[width=0.90\textwidth]{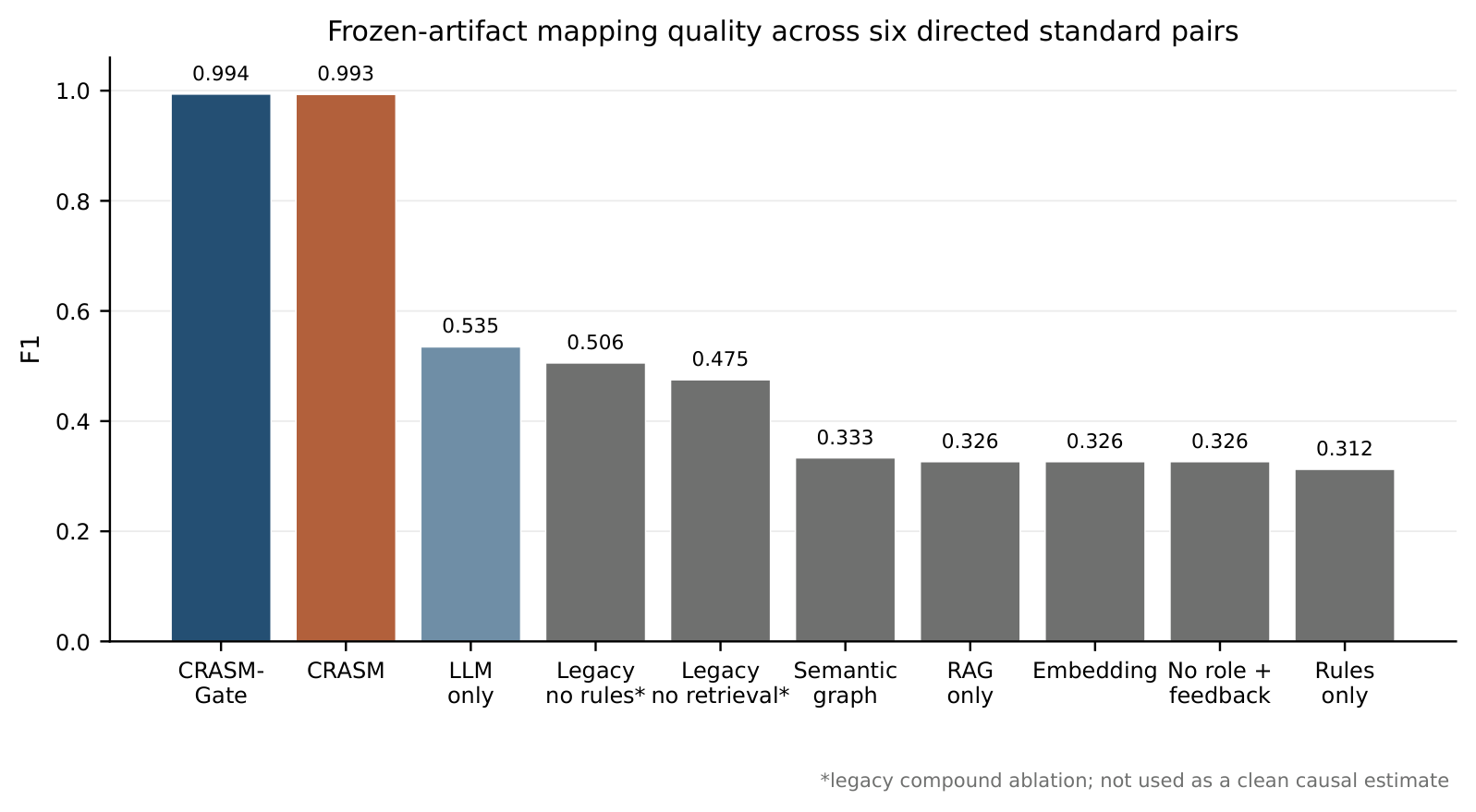}
		\caption{Aggregate F1 by frozen-artifact configuration. CRASM is almost indistinguishable from CRASM-Gate in mapping quality. Legacy compound ablations are shown for provenance but are excluded from isolated component claims.}
		\label{fig:f1}
	\end{figure*}

	Table~\ref{tab:aggregate} and Fig.~\ref{fig:f1} show the frozen-artifact comparison using publication-facing names. CRASM-Gate achieves mean F1 0.9938 across the six directed pairs and deterministic CRASM reaches 0.9931. These correspond to 1431 and 1430 correct top decisions, respectively, out of 1440 samples. Their difference is one decision. LLM-only reaches F1 0.5347 (770/1440), while retrieval-, similarity-, and rule-focused baselines remain near 0.31--0.33.
	
	\begin{table*}[!t]
		\centering
		\caption{Aggregate frozen-artifact results across six directed mapping pairs. SD is descriptive variation across pair-level values, not repeated-seed uncertainty. Rows marked $\dagger$ are legacy compound ablations and are not interpreted as one-component causal estimates.}
		\label{tab:aggregate}
		\small
		\begin{tabular}{lccccr}
			\toprule
			Configuration & F1 & Pair SD & Validity & Observed latency (s/sample) & Correct / 1440 \\
			\midrule
			\textbf{CRASM-Gate} & \textbf{0.9938} & 0.0153 & 1.0000 & 15.7666 & 1431 \\
			\textbf{CRASM} & 0.9931 & 0.0170 & 1.0000 & \textbf{0.0688} & 1430 \\
			LLM-only & 0.5347 & 0.1141 & 1.0000 & 24.5263 & 770 \\
			Legacy w/o Rules$\dagger$ & 0.5056 & 0.1035 & 1.0000 & 11.0734 & 728 \\
			Legacy w/o Retrieval$\dagger$ & 0.4750 & 0.1722 & 0.9993 & 2.0750 & 684 \\
			Semantic Graph Calibrated & 0.3333 & 0.0000 & 1.0000 & 0.0753 & 480 \\
			RAG-only & 0.3264 & 0.0170 & 1.0000 & 0.0672 & 470 \\
			Embedding Similarity & 0.3264 & 0.0170 & 1.0000 & 0.0561 & 470 \\
			Role/Feedback Stack Removed & 0.3264 & 0.0170 & 1.0000 & 0.0639 & 470 \\
			Rules-only & 0.3125 & 0.0510 & 1.0000 & 0.0260 & 450 \\
			\bottomrule
		\end{tabular}
	\end{table*}
	
	The strongest result is therefore not the near-perfect headline score by itself but the decomposition of responsibility: deterministic CRASM provides essentially all measured quality in the controlled setting, while the LLM-only path is far weaker. CRASM-Gate should consequently be interpreted as an optional extension around a strong deterministic core, not as a generative system with deterministic post-processing.
	
	\subsection{Marginal Value of Model Assistance}
	Table~\ref{tab:marginal} makes the CRASM-to-CRASM-Gate increment explicit. The model-assisted framework adds one correct decision over 1440 samples, an absolute F1 increase of approximately $6.94\times10^{-4}$, while the recorded mean latency increases by 15.6978 s/sample. The latency ratio is approximately 229$\times$ in this artifact. Because model/cache activity is heterogeneous, these timings are not hardware-normalized; nevertheless, they show why the model branch must be justified by uncertainty reduction rather than enabled by default.
	
	\begin{table}[!t]
		\centering
		\caption{Observed marginal effect of CRASM-Gate over deterministic CRASM.}
		\label{tab:marginal}
		\small
		\begin{tabular}{lr}
			\toprule
			Quantity & CRASM-Gate $-$ CRASM \\
			\midrule
			F1 & +0.00069 \\
			Correct decisions / 1440 & +1 \\
			Validity & +0.0000 \\
			Latency / sample & +15.6978 s \\
			Latency ratio & 229.3$\times$ \\
			\bottomrule
		\end{tabular}
	\end{table}
	
	The deterministic advantage is also large relative to the model-only baseline. LLM-only makes 670 top-1 errors in the 1440-sample controlled set, whereas CRASM makes 10. Within this frozen artifact, that corresponds to a 98.5\% reduction in top-1 errors and a mean per-sample latency approximately 356\,$\times$ lower (0.0688 s versus 24.5263 s). These ratios are descriptive of the recorded system rather than claims about all LLM deployments, but they reinforce the central result: the decisive information in this task is carried by explicit semantic roles and engineering constraints, not by unconstrained language-model inference.

	\subsection{Error Accounting and Practical Effect Size}
	Aggregate F1 can hide the practical meaning of a small numerical difference near 1.0. Table~\ref{tab:erroraccount} therefore reports the corresponding error counts. LLM-only makes 670 incorrect top-1 decisions, deterministic CRASM makes 10, and CRASM-Gate makes 9. Thus CRASM removes 660 of the 670 LLM-only errors, a 98.5\% reduction in top-1 error count on the controlled artifact. CRASM-Gate removes one additional error relative to CRASM. The latter is a measurable improvement, but its absolute magnitude is one sample and should be interpreted together with the substantially larger observed latency.
	
	\begin{table}[!t]
		\centering
		\caption{Error accounting for the three principal operating points.}
		\label{tab:erroraccount}
		\small
		\begin{tabular}{lrrr}
			\toprule
			Method & Correct & Errors & Latency (s/sample) \\
			\midrule
			CRASM-Gate & 1431 & 9 & 15.7666 \\
			CRASM & 1430 & 10 & 0.0688 \\
			LLM-only & 770 & 670 & 24.5263 \\
			\bottomrule
		\end{tabular}
	\end{table}
	
	The same numbers clarify the systems trade-off. CRASM is approximately $24.5263/0.0688\approx356$ times faster than LLM-only in the recorded end-to-end artifact while achieving substantially higher mapping quality. CRASM-Gate is approximately $15.7666/0.0688\approx229$ times slower than CRASM for one additional correct decision. These ratios are descriptive of the recorded environment, cache state, and model endpoint rather than universal performance constants; they are used to motivate gated profiling rather than to claim hardware-independent speedups.
	
	\subsection{Pair-Level Behavior}
	Table~\ref{tab:pairs} shows the three most informative operating points. Five of six pairs are perfect for both CRASM-Gate and CRASM. All remaining errors are concentrated in AAS$\rightarrow$OPC UA, where CRASM-Gate reaches 0.9625 and CRASM 0.9583. This concentration matters more than the aggregate average: one pair accounts for the complete error budget of both proposed variants.
	
	\begin{table*}[!t]
		\centering
		\caption{Pair-level F1 and observed latency for CRASM-Gate, CRASM, and LLM-only.}
		\label{tab:pairs}
		\scriptsize
		\begin{tabular}{llrr}
			\toprule
			Mapping pair & Configuration & F1 & Latency/sample (s) \\
			\midrule
			\multirow{3}{*}{AAS$\rightarrow$OPC UA} & CRASM-Gate & 0.9625 & 31.0413 \\
			& CRASM & 0.9583 & 0.0965 \\
			& LLM-only & 0.3333 & 31.2947 \\
			\midrule
			\multirow{3}{*}{IEC 61499$\rightarrow$IEEE 1451} & CRASM-Gate & 1.0000 & 37.7348 \\
			& CRASM & 1.0000 & 0.1128 \\
			& LLM-only & 0.6750 & 34.3918 \\
			\midrule
			\multirow{3}{*}{IEEE 1451$\rightarrow$IEC 61499} & CRASM-Gate & 1.0000 & 25.6935 \\
			& CRASM & 1.0000 & 0.0728 \\
			& LLM-only & 0.5583 & 29.2975 \\
			\midrule
			\multirow{3}{*}{ISO15926-style$\rightarrow$AAS} & CRASM-Gate & 1.0000 & 0.0509 \\
			& CRASM & 1.0000 & 0.0500 \\
			& LLM-only & 0.5917 & 17.9113 \\
			\midrule
			\multirow{3}{*}{OPC UA$\rightarrow$AAS} & CRASM-Gate & 1.0000 & 0.0413 \\
			& CRASM & 1.0000 & 0.0417 \\
			& LLM-only & 0.5458 & 16.8034 \\
			\midrule
			\multirow{3}{*}{OPC UA$\rightarrow$IEEE 1451} & CRASM-Gate & 1.0000 & 0.0378 \\
			& CRASM & 1.0000 & 0.0387 \\
			& LLM-only & 0.5042 & 17.4590 \\
			\bottomrule
		\end{tabular}
	\end{table*}
	
	A 0.9938 aggregate score can therefore conceal a direction with nine errors while every other direction is perfect. Deployment should use pair-specific validators, error budgets, and thresholds rather than one global acceptance policy.
	
	\subsection{Difficulty and Contextual-Role Reasoning}
\begin{figure*}[!t]
		\centering
		\includegraphics[width=0.83\textwidth]{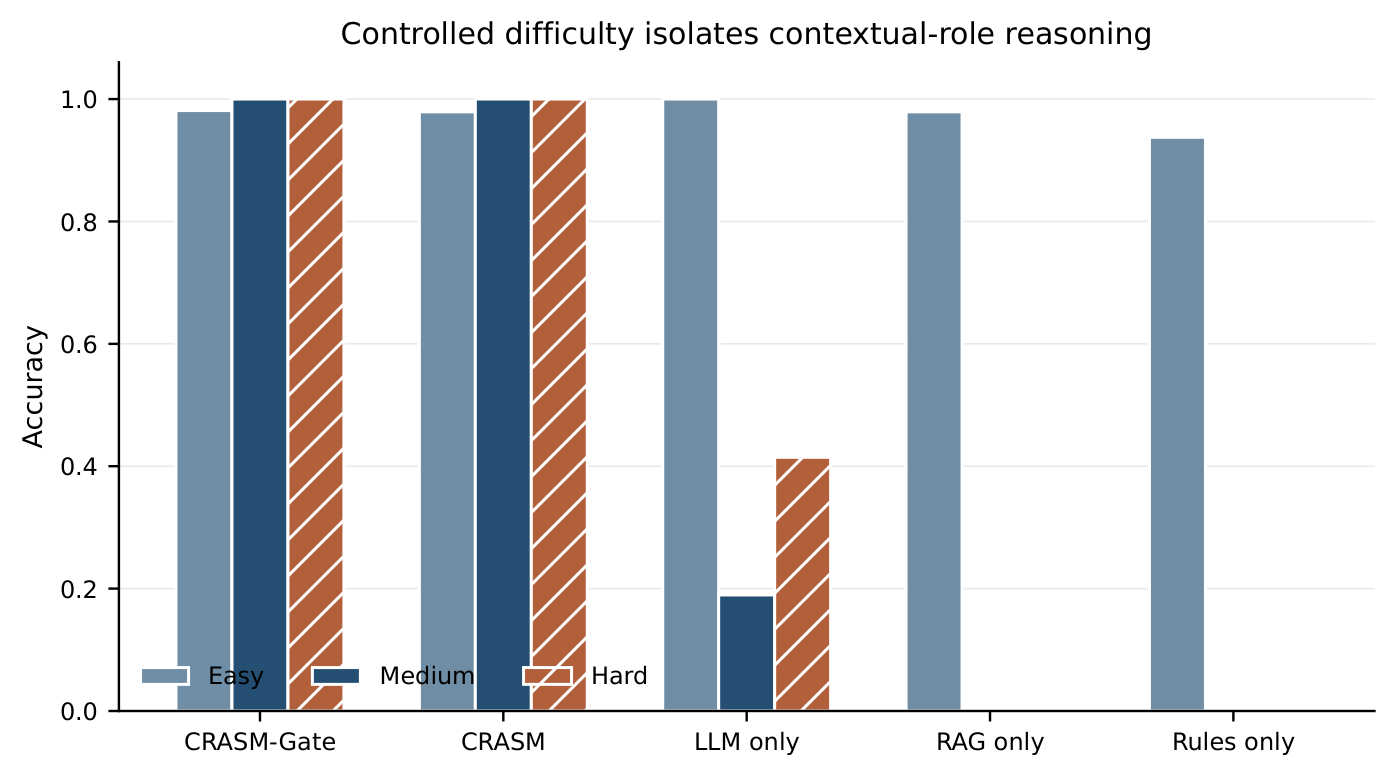}
		\caption{Accuracy by controlled difficulty. Medium and hard samples require contextual-role disambiguation rather than nearest-similarity selection.}
		\label{fig:difficulty}
	\end{figure*}

	Figure~\ref{fig:difficulty} isolates the controlled difficulty effect. CRASM-Gate obtains accuracy 0.9813, 1.0000, and 1.0000 on easy, medium, and hard samples; CRASM obtains 0.9792, 1.0000, and 1.0000. Thus all 960 medium/hard decisions are correct for each proposed method, and their remaining errors occur in the easy AAS$\rightarrow$OPC UA subset.
	
	The baseline pattern is qualitatively different. LLM-only reaches 1.0000 on easy samples but 0.1896 on medium and 0.4146 on hard samples. RAG-only and Rules-only perform well on easy data but reach 0 on both context-conflict tiers. Because medium and hard candidates deliberately share the same signal family, these cases require interpretation of \emph{what the context means}, not merely retrieval of a semantically related target. This is direct mechanism-level evidence for the destination-versus-origin distinction encoded by CRASM.
	
	\subsection{Mechanism Removal and Ablation-Semantics Audit}
	The strongest deterministic mechanism-removal contrast is CRASM versus the frozen \method{ablation\_no\_llm} row. Both operate without generative model calls, but the latter also removes the role-interpretation/feedback stack and reaches F1 0.3264. The gap from 0.9931 to 0.3264 therefore shows that the \emph{combined interpretation/feedback mechanism} is essential in this controlled benchmark; it should not be described as the effect of ``removing the LLM.''
	
	The artifact audit also uncovered a second issue: the frozen no-rules and no-retrieval configurations removed more than the named evidence source because they did not retain the complete role interpreter. Their 0.5056 and 0.4750 values are therefore compound diagnostics, not isolated causal contributions. The current repository has corrected those two definitions by retaining role-aware interpretation. We deliberately do not substitute unexecuted numbers into this manuscript. A fresh run of the cleaned protocol is required before rules-only and retrieval-only component effects are reported causally.
	
	This audit strengthens rather than weakens the core CRASM claim because the principal CRASM-Gate, CRASM, LLM-only, RAG-only, Rules-only, Embedding Similarity, and Semantic Graph comparisons are semantically stable. It also prevents an implementation-labeling error from being amplified into a quantitative scientific claim.
	
	\subsection{Retrieval and Confidence Diagnostics}
	The artifact reports mean retrieval recall of 0.2618 at rank 1 and 0.6597 at rank 5 for configurations using the shared retrieval layer. These values are far below CRASM/CRASM-Gate F1, demonstrating that retrieval is candidate evidence rather than the final decision. Rules, instance/context features, role interpretation, fusion, and validation can recover the correct mapping even when the gold target is not top-ranked by retrieval alone.
	
\begin{figure*}[!t]
		\centering
		\includegraphics[width=0.88\textwidth]{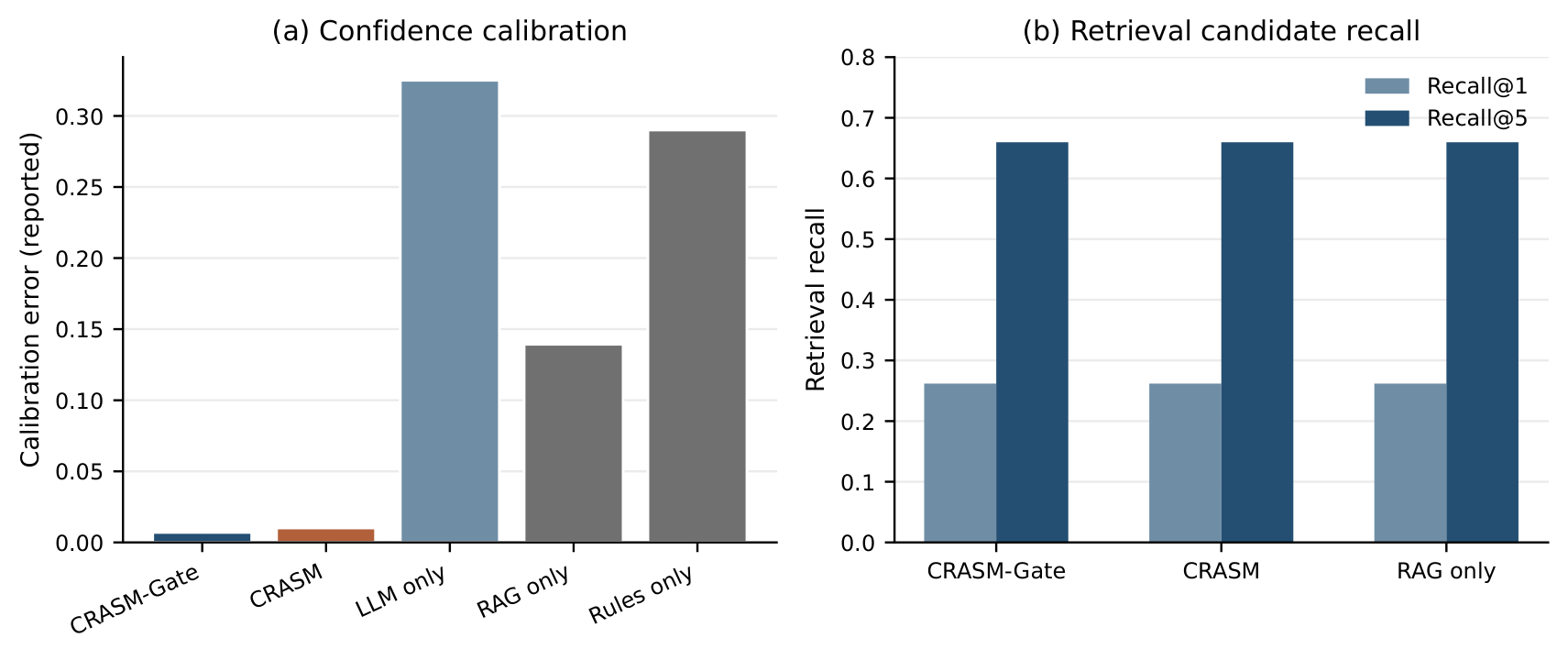}
		\caption{Complementary diagnostics. (a) Reported confidence calibration error. (b) Retrieval recall for configurations sharing the retrieval layer. Retrieval recall is substantially below final CRASM quality, showing that downstream role interpretation and validation materially change the decision.}
		\label{fig:robustness}
	\end{figure*}

	Figure~\ref{fig:robustness} also reports the benchmark confidence-calibration error. CRASM-Gate and CRASM have mean errors of 0.0070 and 0.0102, respectively, while LLM-only has 0.3251. The Semantic Graph Calibrated baseline has an even smaller reported calibration error (0.0018) but only 0.3333 F1. Calibration and discrimination must therefore be interpreted jointly: a model can be numerically well calibrated yet ineffective at selecting the correct target.
	
	\begin{table}[!t]
		\centering
		\caption{Selected diagnostic metrics averaged across six pairs.}
		\label{tab:diagnostics}
		\small
		\begin{tabular}{lccc}
			\toprule
			Configuration & Cal. error & $\Rone$ & $\Rfive$ \\
			\midrule
			CRASM-Gate & 0.0070 & 0.2618 & 0.6597 \\
			CRASM & 0.0102 & 0.2618 & 0.6597 \\
			LLM-only & 0.3251 & 0 & 0 \\
			RAG-only & 0.1394 & 0.2618 & 0.6597 \\
			Rules-only & 0.2901 & 0 & 0 \\
			\bottomrule
		\end{tabular}
	\end{table}
	
	\subsection{Structural Validity and Failure Separation}
	All stable principal configurations in Table~\ref{tab:aggregate} achieve implemented structural-validity rate 1.0000. The legacy no-retrieval row has 0.9993 and contains one recorded low-confidence violation. Structural validity is therefore highly effective at rejecting malformed outputs but does not imply semantic correctness: LLM-only has validity 1.0000 while F1 is 0.5347.
	
	This distinction is central to trustworthy interoperability. Schema-constrained generation and deterministic validation solve the \emph{output-contract} problem; they do not solve the \emph{semantic discrimination} problem. CRASM's role-aware evidence is what separates those two responsibilities in the proposed architecture.
	
	\subsection{Quality--Latency Trade-Off}
	\label{sec:latency}
\begin{figure}[!t]
		\centering
		\includegraphics[width=\columnwidth]{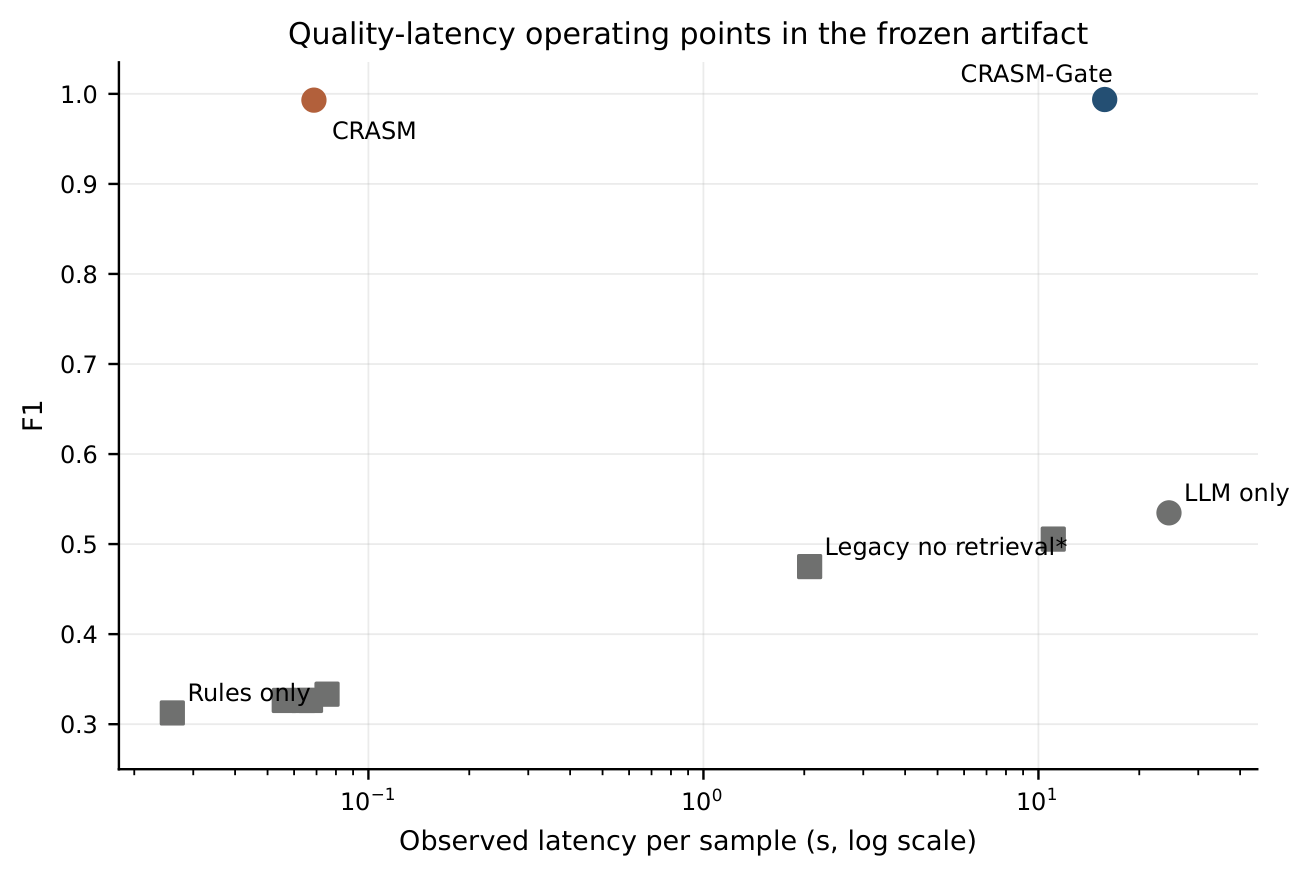}
		\caption{Quality--latency operating points. The horizontal axis is logarithmic. Timings are descriptive artifact measurements, not hardware-normalized inference benchmarks.}
		\label{fig:latency}
	\end{figure}

	Figure~\ref{fig:latency} compares observed mapping quality and latency. CRASM reaches 0.9931 F1 at 0.0688 s/sample, CRASM-Gate reaches 0.9938 at 15.7666 s/sample, and LLM-only reaches 0.5347 at 24.5263 s/sample. In the frozen artifact, the deterministic operating point therefore lies very close to the maximum measured quality while being orders of magnitude cheaper in observed execution time.
	
	The 229$\times$ ratio must be interpreted conservatively because the artifact shows heterogeneous model/token/cache activity across pairs. We therefore do not claim a universal algorithmic speed-up. The stronger systems conclusion is that a model-assisted branch must be \emph{selective and accountable}: on this artifact, dense model involvement buys one additional correct top decision over deterministic CRASM.
	
	\section{Discussion}
	\label{sec:discussion}
	\subsection{The Main Scientific Result Is an Authority Decomposition}
	The central result is not simply that CRASM-Gate reaches 0.9938 F1. Near-perfect scores on a controlled candidate benchmark demand skepticism. The more useful finding is the decomposition of authority: deterministic CRASM reaches 0.9931, solves all controlled medium/hard context-conflict samples, and leaves only a one-decision gap to the model-assisted framework. LLM-only is much weaker, and retrieval/rule baselines fail when candidates share the same signal family but differ in contextual role.
	
	This supports a specific engineering thesis: industrial semantic mapping should be organized around \emph{semantic admissibility}, not nearest similarity. Whether a target is the correct measurement role, belongs to the intended instance, respects units and datatype, and represents destination rather than source provenance are constraints that can be encoded, audited, and tested. A language model may help interpret weakly structured evidence, but the final mapping remains safer when constrained to a candidate vocabulary and independently validated.
	
	\subsection{Why CRASM-Gate Is Not ``CRASM + LLM''}
	The name CRASM-Gate is deliberate. Simply appending an LLM would suggest that model generation is a permanent stage. The intended design is different: CRASM owns normal deterministic decisions; the gate decides whether unresolved evidence merits model assistance; and validation owns final acceptance. This decomposition makes invocation rate, routed-subset quality, latency, and cost independently measurable.
	
	The frozen artifact does not yet demonstrate ideal sparse routing because model activity is recorded broadly. That limitation turns directly into a falsifiable systems question for the next experiment: can the gate preserve CRASM-Gate quality while reducing invocation rate toward the truly ambiguous subset? Reporting that trade-off is more informative than reporting a larger model or a longer prompt.
	
	\subsection{Why Structured Generation Matters but Is Not the Novelty Alone}
	The strict schema removes a subtle source of researcher discretion. Without a typed contract, a model can produce a plausible value such as \method{"confidence": "high"} while the evaluator expects a number. Post-hoc conversion such as ``high $\mapsto$ 0.9'' changes evaluation semantics. Requiring numeric confidence and finite mapping/transform types at generation time avoids that ambiguity and makes schema changes cache-visible.
	
	Yet structured generation alone cannot explain CRASM-Gate quality. LLM-only already satisfies the structural contract but remains far below CRASM. The novel element is therefore the combination of role-aware semantic reasoning, deterministic constraints, selective escalation, and independent validation.
	
	\subsection{Why the Journal Contribution Is Distinct from the Preliminary Paper}
	The preliminary ISynKGR work asked how to benchmark hybrid semantic interoperability reproducibly. This article asks a different question: what decision mechanism actually resolves cross-standard ambiguity, and when does model assistance add enough value to justify its cost? The shift from \emph{benchmark framework} to \emph{mapping method plus gated extension} is substantive. It is reinforced by formal role interpretation, explicit refusal, a typed model boundary, pair/difficulty decomposition, latency economics, and an artifact-semantics audit.
	
	This distinction is important for avoiding the common ``conference paper plus more experiments'' pattern. The journal manuscript does not rely on a larger score as its novelty claim. Its core claim is mechanistic: explicit contextual roles and deterministic structural control account for nearly all measured quality in the controlled setting, while the model is a bounded optional component.
	
	\subsection{Industrial Deployment Implications}
	The results motivate a deterministic-first deployment profile. CRASM is attractive when traceability, repeatability, low latency, and bounded engineering rules dominate. CRASM-Gate is appropriate when descriptions are linguistically complex, metadata are incomplete, or deterministic evidence conflicts. A human-review tier can be placed above the same uncertainty signal so low-margin cases are escalated to a person instead of forced through either deterministic or generative guessing.
	
	Production use still requires controls outside this benchmark: versioned enterprise candidate universes, source/target model versions, executable and unit-tested transforms, companion-specification validators, namespace/reference handling, and pair-specific acceptance policies. The present evidence supports the decision architecture, not unrestricted certification of arbitrary industrial conversions.
	
	\subsection{Why AAS$\rightarrow$OPC UA Remains the Error Concentration}
	All errors of the two proposed methods occur in the easy AAS$\rightarrow$OPC UA subset. ``Easy'' describes the generator's candidate construction, not universal engineering difficulty. The direction still combines asset/submodel semantics with OPC UA object/variable and namespace-like path cues. The correct response is not to tune global weights against nine or ten errors; it is to classify those cases into object-versus-variable, instance, and parent-path failure modes and turn them into targeted regression tests and pair-specific validation rules.

	\section{Deployment, Governance, and Decision Profiles}
	\label{sec:deployment}
	The experimental results suggest two operational profiles rather than one universal configuration. \emph{CRASM} is the natural default when latency, reproducibility, offline execution, and auditability dominate. \emph{CRASM-Gate} is a controlled escalation profile for cases in which deterministic evidence is weak, conflicting, or linguistically incomplete. This distinction is more useful operationally than treating model use as a global on/off feature.
	
	\begin{table*}[!t]
		\centering
		\caption{Recommended operational interpretation of the evaluated configurations.}
		\label{tab:profiles}
		\small
		\begin{tabular}{p{0.15\textwidth}p{0.22\textwidth}p{0.25\textwidth}p{0.27\textwidth}}
			\toprule
			Profile & Intended role & Required observability & Principal caveat \\
			\midrule
			CRASM & Default deterministic mapping & candidate list, feature scores, rules, role interpretation, validator trace & quality depends on available candidates, metadata, and implemented validators \\
			CRASM-Gate & Selective escalation for unresolved cases & all CRASM evidence plus gate reason, model identifier, prompt/schema version, invocation and token/latency record & selectivity and cost must be demonstrated with a dedicated gated run \\
			LLM-only & Experimental baseline & prompt, model output, candidate vocabulary & lower quality in this artifact and insufficient authority controls for deployment \\
			No-match / manual review & Safe fallback when evidence is insufficient & ambiguity reason and rejected candidates & requires downstream review workflow \\
			\bottomrule
		\end{tabular}
	\end{table*}
	
	For governance, the most important distinction is between \emph{explanation} and \emph{authority}. Knowledge graphs and semantic layers can improve explainability \cite{steindl2024semantic,tiddi2022}, and LLMs can produce useful textual rationales, but the final acceptance record should be based on machine-checkable evidence. In CRASM-Gate, model rationale is therefore supplemental provenance. The authoritative record is the candidate vocabulary, deterministic evidence, gate state, schema-valid model contribution if any, and the validator result.
	
	Deployment should also separate threshold configuration from model choice. The confidence floor and ambiguity margin determine which cases are considered unresolved; the selected model only influences the subset routed through that boundary. This separation permits threshold sweeps, model substitution, and offline CRASM operation without changing the semantic-mapping contract. It also makes future comparisons between local/open models and hosted models more interpretable because the deterministic part of the pipeline remains fixed.
	
	Finally, abstention should be exposed to integrating applications rather than hidden as an internal failure. An industrial mapping service can route a no-match or low-margin result to human review, a rule-authoring queue, or a candidate-retrieval expansion step. Treating abstention as a valid operational outcome reduces pressure to convert uncertainty into a plausible but structurally incorrect mapping.
	
	\section{Reproducibility and Artifact Design}
	\label{sec:repro}
	The software project is now named \textbf{CRASM-Gate}. New code uses the public \method{crasm} package and publication-facing runtime modes \method{crasm} and \method{crasm\_gate}. Historical \method{isynkgr} implementation names and scenario IDs are retained only where required to reproduce frozen artifacts and older scripts. This separation avoids allowing implementation history to dictate scientific terminology.
	
	The repository contains standard adapters, canonical models, candidate generation, retrieval, rules, role-aware interpretation, model integration, JSON Schema, validators, scenario orchestration, tests, research statistics, official-source preparation, Docker evaluation, and report generation. The canonical evaluation entry point is '\method{docker compose --profile evaluation up --build}' and defaults to one complete diagnostic run. Publication mode is explicit and supports 20--30 independent replicates, preventing routine development from accidentally launching an expensive model-assisted publication experiment.
	
	The repository also distinguishes runtime and research surfaces: a compact public CRASM API/CLI for translation, a benchmark layer for controlled experiments, and an evaluation container for end-to-end scientific checks. Machine-readable results carry stable internal IDs together with publication-facing CRASM labels. This makes naming migrations auditable while preserving reproducibility.
	
	The frozen result files supplied with this manuscript remain the source of the numerical tables and figures. They are not silently rewritten to imitate the corrected ablation semantics. For peer review, the recommended package is therefore two-part: (i) the immutable frozen result snapshot used by this article and (ii) the current code release that documents the corrected scenario contract. A final archival release should additionally record the exact commit, environment, model identifier, and publication-run seed manifest.
	
	\section{Conclusion}
	\label{sec:conclusion}
	This article introduced \crasm, a deterministic constraint- and role-aware semantic mapping method, and \crasmgate, its selectively model-assisted extension for candidate-constrained interoperability across heterogeneous industrial standards. The architecture separates canonicalization, retrieval, deterministic rules, destination-versus-origin role interpretation, semantic/structural ranking, optional bounded model evidence, and final deterministic validation. The method therefore makes an explicit distinction between \emph{semantic similarity} and \emph{structural/role admissibility}.
	
	Across the frozen 14,400-decision artifact, CRASM-Gate reaches mean F1 0.9938 and implemented structural-validity rate 1.0000; deterministic CRASM reaches 0.9931 without generative-model calls; and LLM-only reaches 0.5347. CRASM and CRASM-Gate are perfect on all controlled medium and hard context-conflict samples. The model-assisted framework improves the deterministic core by only one correct decision out of 1440 while incurring substantially larger observed latency in this artifact. The narrow conclusion is therefore strong and auditable: explicit contextual-role interpretation and deterministic structural control explain nearly all measured quality, while model assistance is best treated as a gated uncertainty resolver rather than workflow authority.
	
	The artifact audit further shows why experimental semantics matter. Legacy compound ablations should not be converted into causal component claims after a refactor. The next publication-grade step is a fresh multi-seed run of the cleaned ablation protocol, followed by expert-annotated official industrial artifacts, external ontology-matching baselines, unrestricted candidate discovery, and executable cross-model transformations with pair-specific validators. Those experiments test whether CRASM's mechanism survives outside the controlled environment rather than merely increasing the size of the benchmark.
	
	\section*{Acknowledgment}
	The authors acknowledge the support of the EU-funded Arrowhead fPVN project under KDT Joint Undertaking Grant Agreement No. 101111977.


\begin{thebibliography}{99}
		
		\bibitem{barbu2024}
		M.~Barbu, A.~V.~Vevera, and D.~C.~Barbu, ``Standardization and interoperability--key elements of digital transformation,'' in \emph{Digital Transformation: Technology, Tools, and Studies}, Cham, Switzerland: Springer Nature Switzerland, 2024, pp.~87--94.
		
		\bibitem{petrasch2022}
		R.~J.~Petrasch and R.~R.~Petrasch, ``Data integration and interoperability: Towards a model-driven and pattern-oriented approach,'' \emph{Modelling}, vol.~3, no.~1, pp.~105--126, 2022, doi: 10.3390/modelling3010008.
		
		\bibitem{zeid2019}
		A.~Zeid, S.~Sundaram, M.~Moghaddam, S.~Kamarthi, and T.~Marion, ``Interoperability in smart manufacturing: Research challenges,'' \emph{Machines}, vol.~7, no.~2, p.~21, 2019, doi: 10.3390/machines7020021.
		
		\bibitem{perzylo2019}
		A.~Perzylo, S.~Profanter, M.~Rickert, and A.~Knoll, ``OPC UA NodeSet ontologies as a pillar of representing semantic digital twins of manufacturing resources,'' in \emph{Proc. 24th IEEE Int. Conf. Emerging Technologies and Factory Automation (ETFA)}, 2019, pp.~1085--1092.
		
		\bibitem{ochoa2023}
		W.~Ochoa, F.~Larrinaga, and A.~Perez, ``Architecture for managing AAS-based business processes,'' \emph{Procedia Computer Science}, vol.~217, pp.~217--226, 2023, doi: 10.1016/j.procs.2022.12.217.
		
		\bibitem{jimenezruiz2011}
		E.~Jimenez-Ruiz and B.~Cuenca Grau, ``LogMap: Logic-based and scalable ontology matching,'' in \emph{The Semantic Web--ISWC 2011}, Springer, 2011, pp.~273--288.
		
		\bibitem{faria2013}
		D.~Faria, C.~Pesquita, E.~Santos, M.~Palmonari, I.~F.~Cruz, and F.~M.~Couto, ``The AgreementMakerLight ontology matching system,'' in \emph{On the Move to Meaningful Internet Systems: OTM 2013 Conferences}, Springer, 2013, pp.~527--541.
		
		\bibitem{ji2021}
		S.~Ji, S.~Pan, E.~Cambria, P.~Marttinen, and P.~S.~Yu, ``A survey on knowledge graphs: Representation, acquisition, and applications,'' \emph{IEEE Trans. Neural Netw. Learn. Syst.}, vol.~33, no.~2, pp.~494--514, 2021.
		
		\bibitem{tiddi2022}
		I.~Tiddi and S.~Schlobach, ``Knowledge graphs as tools for explainable machine learning: A survey,'' \emph{Artificial Intelligence}, vol.~302, p.~103627, 2022.
		
		\bibitem{giglou2023}
		H.~Babaei Giglou, J.~D'Souza, and S.~Auer, ``LLMs4OL: Large language models for ontology learning,'' in \emph{The Semantic Web--ISWC 2023}, Springer, 2023, pp.~408--427.
		
		\bibitem{giglou2024llms4om}
		H.~Babaei Giglou, J.~D'Souza, F.~Engel, and S.~Auer, ``LLMs4OM: Matching ontologies with large language models,'' in \emph{The Semantic Web: ESWC 2024 Satellite Events, Part I}, Springer, 2024, pp.~25--35, doi: 10.1007/978-3-031-78952-6\_3.
		
		\bibitem{tufek2024validating}
		N.~Tufek, A.~S.~Thuluva, V.~P.~Just, F.~J.~Ekaputra, T.~Bandyopadhyay, M.~Sabou, and A.~Hanbury, ``Validating semantic artifacts with large language models,'' in \emph{The Semantic Web: ESWC 2024 Satellite Events, Part I}, Springer, 2024, pp.~92--101, doi: 10.1007/978-3-031-78952-6\_9.
		
		\bibitem{mohsenzadegan2026isynkgr}
		K.~Mohsenzadegan, V.~Tavakkoli, and K.~Kyamakya, ``ISynKGR: An adaptive benchmark framework for cross-standard semantic interoperability,'' in \emph{Int. Workshop on AI \& Mathematical Methods for Real-world Impact (AI2M4RI)}, Athens, Greece, Aug.~18--20, 2026.
		
		\bibitem{darocha2020}
		H.~Da Rocha, A.~Espirito-Santo, and R.~Abrishambaf, ``Semantic interoperability in the industry 4.0 using the IEEE 1451 standard,'' in \emph{Proc. 46th Annual Conf. IEEE Industrial Electronics Society (IECON)}, 2020, pp.~5243--5248, doi: 10.1109/IECON43393.2020.9254274.
		
		\bibitem{prenzel2019}
		L.~Prenzel, A.~Zoitl, and J.~Provost, ``IEC 61499 runtime environments: A state-of-the-art comparison,'' in \emph{Computer Aided Systems Theory--EUROCAST 2019}, Springer, 2019, pp.~234--241.
		
		\bibitem{kim2022}
		B.~Kim, S.~Kim, H.~Teijgeler, J.~Lee, J.~Y.~Lee, D.~Lim, H.~W.~Suh, and D.~Mun, ``Use of asset administration shell coupled with ISO 15926 to facilitate the exchange of equipment condition and health status data of a process plant,'' \emph{Applied Sciences}, vol.~12, no.~4, p.~2155, 2022, doi: 10.3390/app12042155.
		
		\bibitem{steindl2024semantic}
		G.~Steindl, T.~Schwarzinger, K.~Schreiberhuber, and F.~J.~Ekaputra, ``Toward semantic event-handling for building explainable cyber-physical systems,'' \emph{IEEE Open Journal of the Industrial Electronics Society}, vol.~5, pp.~928--945, 2024, doi: 10.1109/OJIES.2024.3447001.
		
		\bibitem{maedche2001}
		A.~Maedche and S.~Staab, ``Ontology learning for the semantic web,'' \emph{IEEE Intelligent Systems}, vol.~16, no.~2, pp.~72--79, 2001.
		
		\bibitem{pan2024unifying}
		S.~Pan, L.~Luo, Y.~Wang, C.~Chen, J.~Wang, and X.~Wu, ``Unifying large language models and knowledge graphs: A roadmap,'' \emph{IEEE Trans. Knowl. Data Eng.}, vol.~36, no.~7, pp.~3580--3599, 2024.
		
		\bibitem{yao2023react}
		S.~Yao, J.~Zhao, D.~Yu, N.~Du, I.~Shafran, K.~Narasimhan, and Y.~Cao, ``ReAct: Synergizing reasoning and acting in language models,'' in \emph{Int. Conf. Learning Representations (ICLR)}, 2023.
		
		\bibitem{geng2023grammar}
		S.~Geng, M.~Josifoski, M.~Peyrard, and R.~West, ``Grammar-constrained decoding for structured NLP tasks without finetuning,'' in \emph{Proc. 2023 Conf. on Empirical Methods in Natural Language Processing (EMNLP)}, Association for Computational Linguistics, 2023, pp.~10932--10952, doi: 10.18653/v1/2023.emnlp-main.674.
		
	\end{thebibliography}
\end{document}